\documentclass[letterpaper,10pt,conference]{ieeeconf}

\IEEEoverridecommandlockouts
\usepackage{times}
\usepackage{amsmath}
\usepackage{amssymb}
\usepackage{graphicx}
\usepackage{booktabs} 
\usepackage{array}
\usepackage[table]{xcolor}
\usepackage{cite}
\usepackage{url}
\usepackage{xspace}
\usepackage[normalem]{ulem}
\makeatletter
\let\NAT@parse\undefined
\makeatother
\usepackage[hidelinks]{hyperref}

\newcommand{\method}{\textsc{HACo}\xspace}

\definecolor{ablationdrop}{HTML}{809361}
\newcommand{\abldrop}[1]{\textcolor{ablationdrop}{(-#1\%)}}

\title{\LARGE \bf
HACo: Learning Haptic Active Compliance for \\
Force-Aware Dexterous Manipulation
}
\newif\ifanonymoussubmission
\anonymoussubmissionfalse
\ifanonymoussubmission
\author{Anonymous Authors}
\else
\author{Naisheng Ye$^{1,2,\dagger}$, Yinzhe Zhou$^{2,3}$,
Junkai Zhao$^{2,*}$, Yuhang Lu$^{1,2}$,
Checheng Yu$^{1}$ \\ Zhenjie Yang$^{1}$, Pengwei Wang$^{2}$
and Hongyang Li$^{1}$%
\\[2mm]
\url{https://opendrivelab.github.io/Haco-Page/}
\thanks{$^{1}$The University of Hong Kong, Hong Kong SAR, China;
$^{2}$Beijing Academy of Artificial Intelligence (BAAI), Beijing, China;
$^{3}$Johns Hopkins University, Baltimore, MD, USA.}%
\thanks{$^{\dagger}$Work done during an internship at BAAI.}%
\thanks{$^{*}$Project leader.}}
\fi

\begin{document}

\IEEEaftertitletext{%
  \begin{center}
    \includegraphics[width=\textwidth]{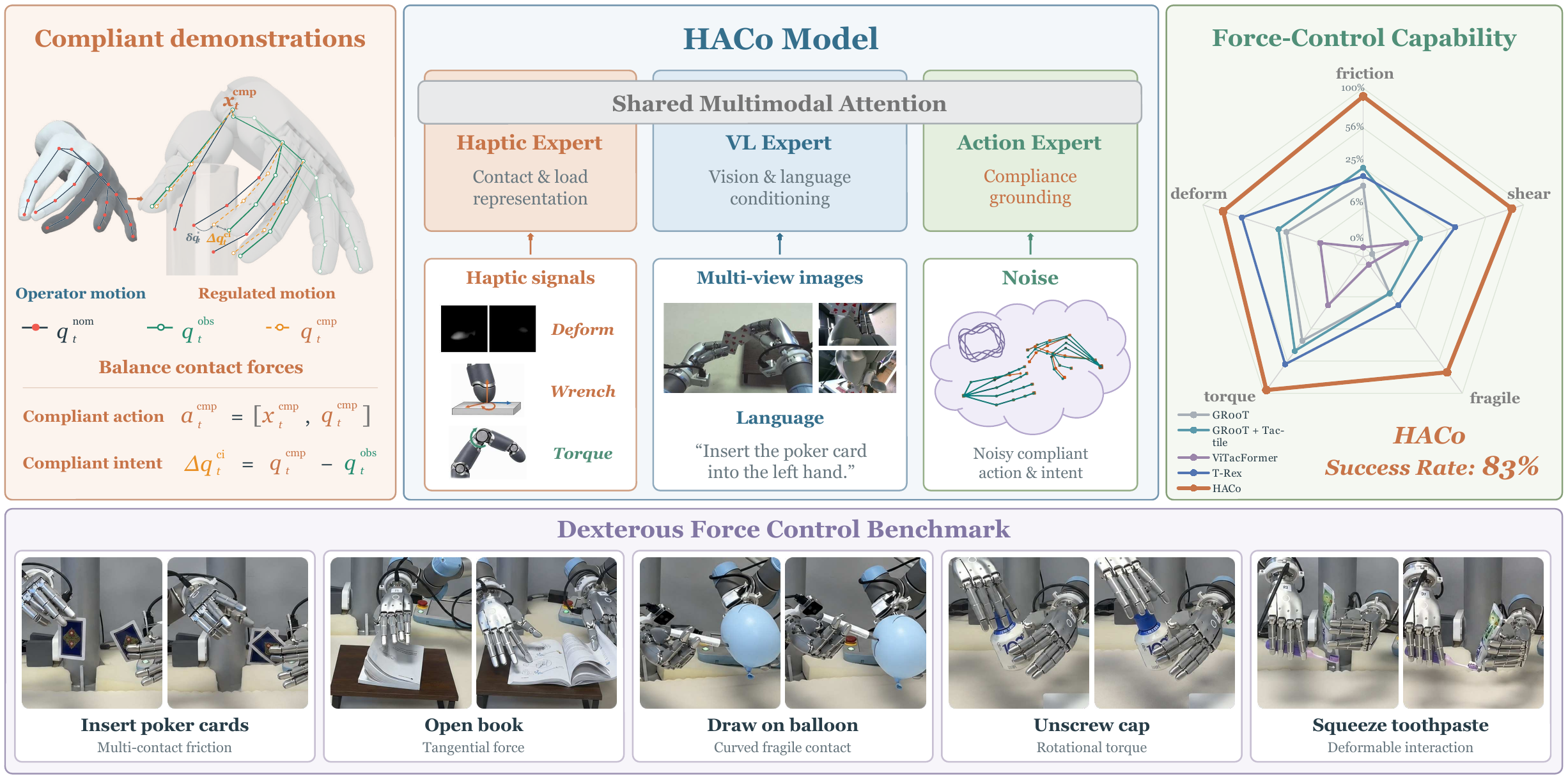}\\[3pt]
    \refstepcounter{figure}\label{fig:teaser}%
    \begin{minipage}{0.94\textwidth}
      \footnotesize
      Fig.~\thefigure.\quad\textbf{Learning haptic active compliance for
      dexterous manipulation.} Compliance-regulated teleoperation produces
      compliant action targets. The discrepancy between compliant hand commands
      and observed states provides compliant-intent supervision. HACo learns these actions
      from visual, language, and haptic observations, with a Compliance Grounding Module
      integrating complementary fingertip tactile and joint-torque features into
      action generation.
      Five real-world tasks evaluate
      manipulation under multi-contact friction, tangential force, fragile
      contact, rotational torque, and object deformation.
    \end{minipage}
  \end{center}
  \vspace{2pt}%
}

\maketitle
\thispagestyle{empty}
\pagestyle{empty}

\begin{abstract}
Contact-rich dexterous manipulation requires policies to translate physical
feedback into motion commands that regulate interaction loads across evolving
multi-contact interactions. This requires both haptic observations that capture
the contact state and action supervision that demonstrates how motion commands
should adapt to it. Existing policies often overlook the complementary roles of
fingertip tactile sensing and joint torque. Meanwhile, commonly used action
targets include nominal teleoperation commands, which can encode excessive
loading, and observed configurations, which omit motion constrained by the
object.
We introduce \textbf{HACo}, a \textbf{\uline{H}aptic \uline{A}ctive \uline{Co}mpliance} policy that learns force-regulating actions
directly from the haptic feedback. 
Compliance-regulated teleoperation converts
operator commands into controller-executable compliant actions that preserve motion intent while regulating interaction loads. HACo learns these compliant
actions directly, with their command--state discrepancy providing auxiliary compliant-intent supervision.
HACo further combines local contact responses from fingertip tactile sensing
with load transmission through the articulated hand captured by joint-torque
feedback, including contacts beyond tactile coverage.
A \textbf{Compliance Grounding
Module} grounds action generation in the evolving haptic state through gated haptic cross-attention, enabling closed-loop force regulation without explicit
online contact modeling. We evaluate HACo on a real-world benchmark spanning multi-contact friction, tangential interaction, fragile curved-surface contact,
rotational torque, and deformable-object manipulation. Across 20 trials per
task, HACo achieves an 83\% mean success rate versus 35\% for the strongest evaluated
baseline. These results demonstrate HACo's ability to translate complementary
haptic feedback into active compliance across diverse force-sensitive
dexterous manipulation tasks.
\end{abstract}

\section{Introduction}
\label{sec:introduction}

Contact-rich dexterous manipulation requires control over both motion and
interaction force. As contacts change across the hand, a geometrically correct
trajectory can fail through excessive pressure, insufficient traction, or
poorly regulated torque. These visually ambiguous failures motivate a closed
loop from physical feedback to force-regulating actions.

Tactile-augmented policies improve contact awareness through cross-modal
alignment~\cite{heng2025vitacformer,huang2025tactilevla,zhang2025vtla,cheng2025omnivtla},
future-interaction prediction
\cite{wu2026tactilewam,huang2026vitacworld,ma2026feelworld}, and
haptic-conditioned action generation
\cite{niu2026trex,yuan2026ftp,li2026fmvla}.
When direct tactile sensing is unavailable or incomplete, joint torque and
related proprioceptive signals are often used as proxies for interaction loads
\cite{yu2025forcevla,zhang2025tavla,ma2026current,wong2026implicit,zhao2026crg}.
However, the complementary roles of fingertip tactile sensing and joint torque
remain underexplored in existing policies. Touch resolves local deformation,
shear, and slip cues, while torque reflects loads transmitted through the hand,
including contacts beyond tactile coverage.

Beyond sensing contact, a policy must learn how to adjust its actions in
response.
Existing methods typically learn observed robot configurations or nominal
teleoperation commands. Under sustained contact, observed configurations capture
only realized motion, omitting the command--state discrepancy needed to sustain
interaction force. Nominal commands preserve the operator's motion intent,
but may induce excessive force because the operator does not directly perceive
the robot's contact loads.

Compliance control provides a means of regulating interaction forces through
motion references. Impedance and admittance control shape the relation between
motion and force~\cite{hogan1985impedance,keemink2018admittance}, while hybrid
position--force control regulates them along selected task-space
directions~\cite{mason1981compliance,raibert1981hybrid}.
Their reliance on contact assumptions, prescribed control parameters, and
analytical force--motion mappings limits scalability to dexterous hands
with high-dimensional kinematic coupling and changing multi-point contacts.

We introduce \textbf{HACo}, a \textbf{Haptic Active Compliance} policy whose
overall framework is illustrated in Figure~\ref{fig:teaser}. Our key insight is
to use compliance regulation to construct action supervision. During data
collection, arm admittance and
contact-aware hand adjustment transform the operator's nominal references
into controller-executable \emph{compliant actions}. These actions retain
task-directed motion while reducing excessive interaction loads. HACo learns
them directly, with their discrepancy from observed hand state providing
auxiliary \emph{compliant-intent supervision}.

To condition these actions on contact, HACo couples fingertip tactile and
joint-torque feedback along the hand's kinematic chains to form structured
haptic features. Naively concatenating these features with visual or action
tokens may disrupt pretrained representations and does not explicitly connect
contact feedback to action generation. A \emph{Compliance Grounding Module}
(CGM) makes this connection through gated haptic cross-attention, allowing
action features to query the haptic representation as actions are generated.
This formulation enables HACo to regulate interaction loads through
haptic-conditioned motion references, without explicit force prediction,
force tracking, or online analytical contact modeling.

We evaluate HACo on five real-world dexterous tasks spanning multi-contact
friction, tangential interaction, fragile curved-surface contact, rotational
torque, and deformable-object manipulation. HACo achieves an 83\% mean success
rate versus 35\% for the strongest evaluated baseline. Our contributions are
threefold:
\begin{itemize}
    \item We develop complementary haptic perception that couples fingertip
    tactile sensing with joint-torque feedback to represent both local contact
    and loads transmitted through the articulated hand.
    \item We formulate active compliance learning from regulated
    demonstrations, combining controller-executable compliant actions,
    compliant-intent supervision, and the Compliance Grounding Module for
    haptic-conditioned action generation.
    \item We introduce a real-world dexterous force benchmark spanning
    multi-contact friction, tangential interaction, fragile curved-surface
    contact, rotational torque, and deformable-object manipulation.
\end{itemize}

\section{Related Work}
\label{sec:related-work}

\subsection{Haptic Perception for Dexterous Manipulation}

Tactile sensing has been widely studied for robot manipulation, with prior work
exploring tactile representation learning and visuotactile fusion
\cite{lin2024learning,wu2025canonical,sferrazza2023power,fu2024tvl,guzey2023dexterity}.
Recent policies incorporate touch through cross-modal alignment
\cite{heng2025vitacformer,huang2025tactilevla,zhang2025vtla,cheng2025omnivtla,yuan2026ftp},
future interaction prediction
\cite{wu2026tactilewam,huang2026vitacworld,ma2026feelworld}, or reactive action
refinement~\cite{xue2025reactive,niu2026trex}. Beyond fingertip measurements,
force and joint-torque sensing, together with proprioceptive proxies such as
motor current and tracking error, have been investigated to expose interaction loads
\cite{yu2025forcevla,zhang2025tavla,ma2026current,wong2026implicit,zhao2026crg},
while contact wrench and compliance priors have supported skill learning and
grasp synthesis~\cite{zhu2026chord,chen2024springgrasp}. These approaches motivate
the use of complementary physical observations. HACo combines fingertip
tactile and joint-torque feedback as complementary haptic observations, covering
both local contact responses and loads transmitted through the articulated hand.

\subsection{Compliance Control for Contact-Rich Manipulation}

Compliance control broadly denotes interaction-control strategies that shape
the relationship between robot motion and environmental force. Two common
realizations differ in their causality: impedance control produces force in response to motion deviation,
whereas admittance control converts measured force into a motion reference for
a lower-level tracking controller
\cite{hogan1985impedance,keemink2018admittance}. Hybrid position--force control
instead regulates motion and force along selected task-space directions
\cite{mason1981compliance,raibert1981hybrid}. Subsequent work learns or adapts
compliance profiles, impedance parameters, and force--motion residuals while
retaining an explicit low-level control formulation
\cite{kronander2014learning,martin2019variable,hou2025adaptive,shi2026minimalist,ge2025filic}.
Force-aware data collection further applies compliant corrections during
teleoperation or human intervention
\cite{liu2024forcemimic,liu2026dexteleop0,choi2026wild,xu2025crdagger}.
Contact-Grounded Policy~\cite{xu2026cgp} predicts robot state and tactile
feedback and uses a learned contact-consistency mapping to obtain executable
compliance-controller targets. HACo shares the goal of grounding contact in
controller-executable actions, but learns regulated references directly and
uses their discrepancy from observed hand state as auxiliary supervision.
This distinction concerns the supervision and prediction pathway, without
requiring an intermediate prediction of future tactile outcomes.

\begin{figure*}[!t]
  \centering
  \includegraphics[width=0.98\textwidth]{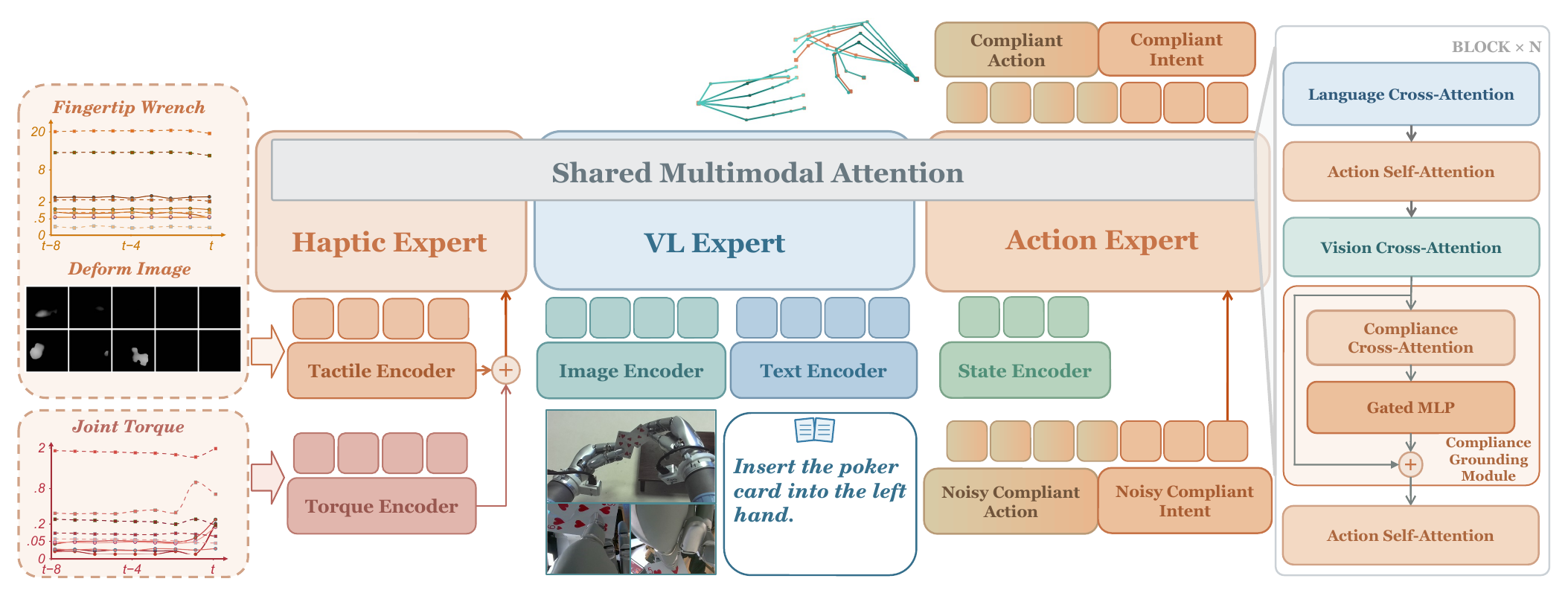}
  \caption{\textbf{Overview of HACo.} The haptic expert aligns fingertip
  wrench and deformation features with joint-torque features from the same
  finger before cross-finger fusion. The Compliance Grounding Module (right)
  lets action features query this haptic representation through gated haptic
  cross-attention alongside vision-language conditioning. Trained with
  conditional flow matching, the action expert learns compliant actions from
  force-regulated demonstrations, with auxiliary compliant-intent supervision
  capturing the discrepancy between compliant hand commands and observed
  hand states.}
  \label{fig:haco-overview}
\end{figure*}

\section{Method}
\label{sec:method}

At time step $t$, HACo receives multi-view RGB observations $\mathbf I_t$, a
language instruction $\ell$, robot state $\mathbf s_t$, and short histories of
fingertip tactile measurements $\mathcal T_t$ and hand-joint torques
$\boldsymbol\tau_t$. It predicts a horizon-$H$ chunk of controller-executable
\emph{compliant actions},
\begin{equation}
\begin{aligned}
\mathcal O_t
  &=\{\mathbf I_t,\ell,\mathbf s_t,
      \mathcal T_{t-L+1:t},\boldsymbol\tau_{t-L+1:t}\},\\
\widehat{\mathbf A}^{\rm cmp}_{t:t+H-1}
  &=\pi_\theta(\mathcal O_t), \qquad
\mathbf a_t^{\rm cmp}=[\mathbf x_t^{\rm cmp},\mathbf q_t^{\rm cmp}],
\end{aligned}
\label{eq:haco_problem}
\end{equation}
where $\mathbf x^{\rm cmp}$ and $\mathbf q^{\rm cmp}$ are the arm end-effector
and dexterous-hand references, respectively. Both are sent directly to the
corresponding low-level controllers. During training, the policy additionally
predicts a compliant intent for the hand, $\Delta\mathbf q^{\rm ci}$, which provides
auxiliary supervision but is never added to the executed action.
Figure~\ref{fig:haco-overview} summarizes HACo's overall architecture. HACo
represents compliance in the motion-reference space, where the Compliance
Grounding Module conditions the flow-based action expert on haptic observations
of interaction loads, enabling it to predict compliant actions rather than
prescribed force targets.

\subsection{Compliance-Regulated Teleoperation}
\label{sec:method_teleoperation}

\begin{figure}[t]
  \centering
  \includegraphics[width=\columnwidth]{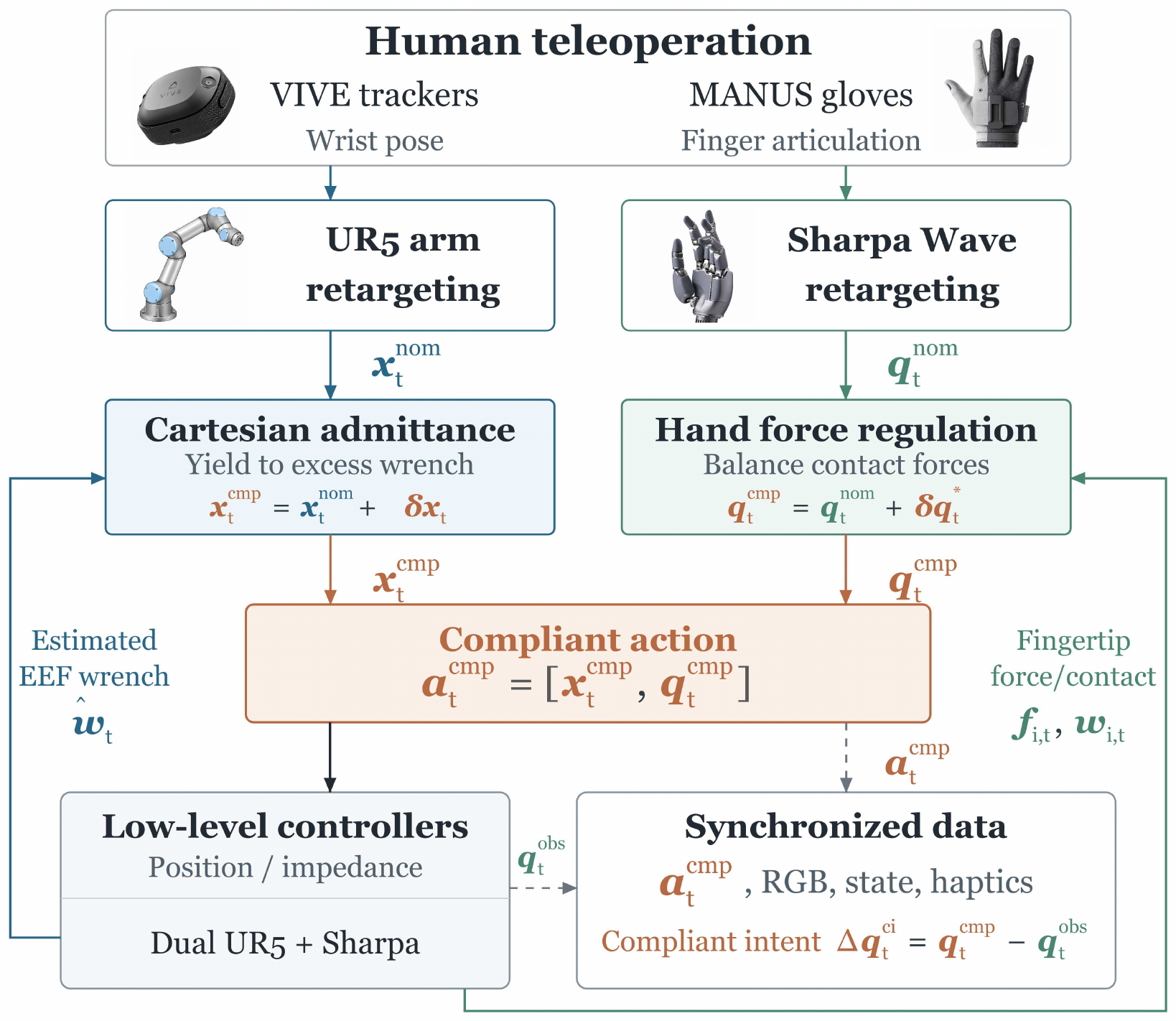}
  \caption{\textbf{Compliance-regulated teleoperation.} Motion retargeting
  provides nominal arm and hand references. Cartesian admittance and
  hand force regulation convert them into controller-executable
  compliant references while synchronized robot state and haptic observations
  are recorded.}
  \label{fig:compliant-teleoperation}
\end{figure}

As illustrated in Figure~\ref{fig:compliant-teleoperation}, our robot runs
low-level position/impedance controllers, while an outer-loop regulator modifies
the teleoperated references before execution. This design
retains the operator's task-directed motion while preventing demonstrations
from encoding unnecessarily large interaction loads. Let
$\mathbf x_t^{\rm nom}$ and $\mathbf q_t^{\rm nom}$ denote the nominal arm and
hand references produced by motion retargeting.

For the arm, the components of the estimated end-effector wrench
$\widehat{\mathbf w}_t$ that exceed an admissible range drive a Cartesian
admittance response. Denoting the excess wrench by
$\mathbf r_w(\widehat{\mathbf w}_t)$, the reference displacement
$\delta\mathbf x_t$ follows a virtual mass--damping--stiffness system:
\begin{equation}
\begin{aligned}
\mathbf M_a\delta\ddot{\mathbf x}_t
 &=-\mathbf D_a\delta\dot{\mathbf x}_t
   -\mathbf K_a\delta\mathbf x_t
   -\mathbf r_w(\widehat{\mathbf w}_t),\\
\mathbf x_t^{\rm cmp}
 &=\mathbf x_t^{\rm nom}+\delta\mathbf x_t.
\end{aligned}
\label{eq:arm_admittance}
\end{equation}
Here, $\mathbf M_a$, $\mathbf D_a$, and $\mathbf K_a$ specify the virtual
admittance dynamics. The excess-wrench operator $\mathbf r_w$ retains only the
amount outside the admissible range. Inside that range, its input vanishes and
the displacement relaxes toward zero under the virtual dynamics.

For the hand, each active fingertip force $\mathbf f_{i,t}$ is clipped in
magnitude to $\bar{\mathbf f}_{i,t}$ without changing its direction. A virtual
contact stiffness $\mathbf K_i$ maps the required unloading to a fingertip
displacement
$\mathbf d_{i,t}=\mathbf K_i^{-1}(\bar{\mathbf f}_{i,t}-\mathbf f_{i,t})$.
We obtain a joint-space reference adjustment through
\begin{equation}
\begin{aligned}
\delta\mathbf q_t^*
 &=\arg\min_{\delta\mathbf q}
 \sum_i w_{i,t}\|\mathbf J_{i,t}\delta\mathbf q-\mathbf d_{i,t}\|_2^2\\[-1mm]
 &\hspace{8mm}+\lambda_r\|\delta\mathbf q\|_2^2\\[-1mm]
 &\hspace{8mm}+\lambda_s
 \|\delta\mathbf q-\delta\mathbf q_{t-1}^*\|_2^2,\\
\mathbf q_t^{\rm cmp}
 &=\mathbf q_t^{\rm nom}+\delta\mathbf q_t^*.
\end{aligned}
\label{eq:hand_compliance}
\end{equation}
Here, $\mathbf J_{i,t}$ is the fingertip contact Jacobian and
$w_{i,t}\in[0,1]$ smoothly activates the corresponding objective upon contact.
The regularizers limit unnecessary joint motion and promote temporal
continuity, and joint-position and update limits are imposed during
optimization. The objective balances fingertip unloading against small, smooth
joint adjustments; kinematic coupling can also move other fingertips.
Here $\mathbf K_i$ is a virtual gain rather than an identified object stiffness.
We record
$\mathbf a_t^{\rm cmp}=[\mathbf x_t^{\rm cmp},\mathbf q_t^{\rm cmp}]$ as the
compliant action ground truth.

\subsection{Active Compliance Learning}
\label{sec:method_active_compliance}

\begin{figure}[t]
  \centering
  \includegraphics[width=\columnwidth]{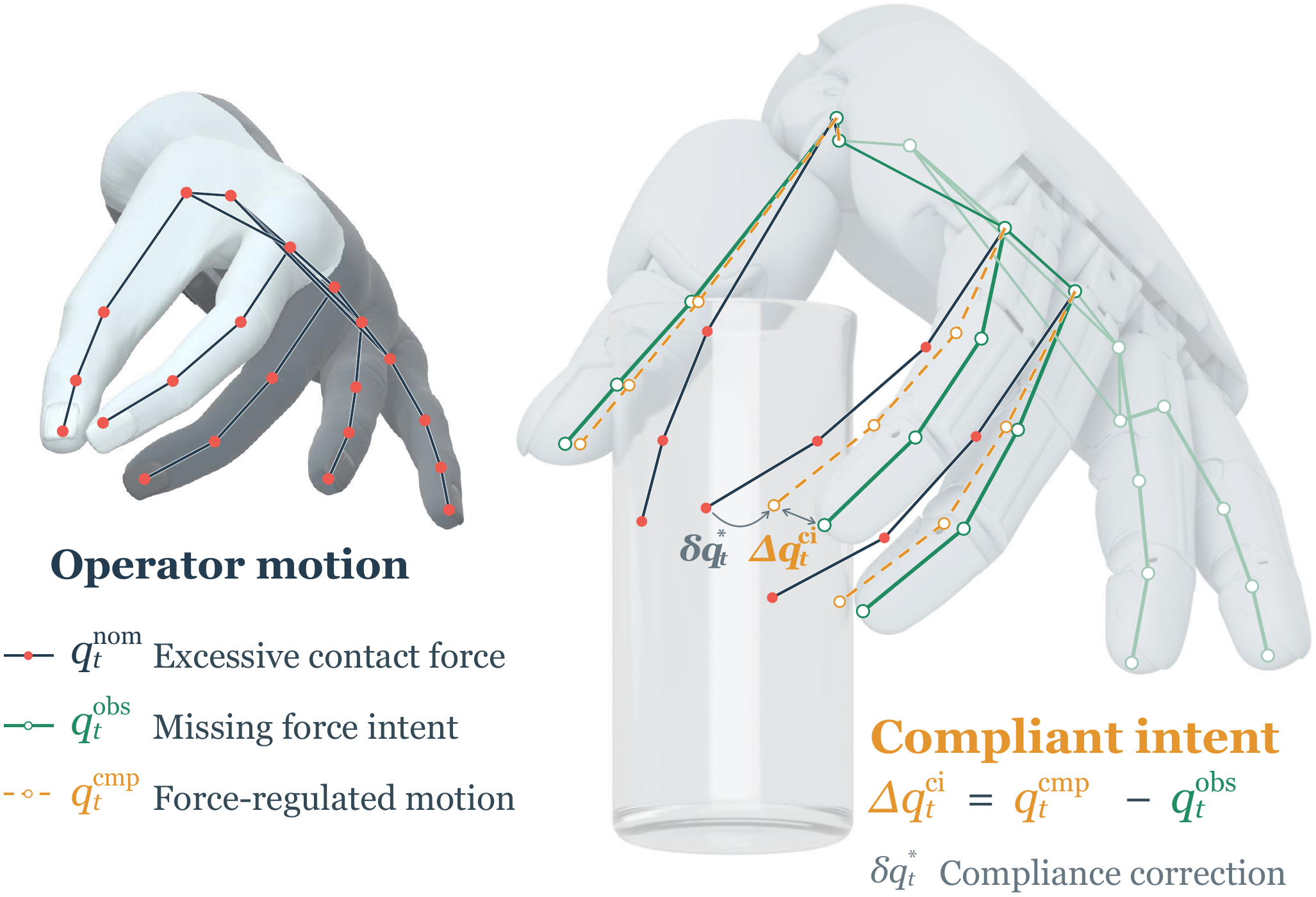}
  \caption{\textbf{Hand action semantics under contact.} Nominal retargeting
  can command excessive contact force, while observed configurations alone
  omit motion constrained by contact. The compliant command preserves
  task-directed motion with force regulation. Its discrepancy from the
  observed state, $\Delta\mathbf q_t^{\rm ci}$, provides compliant-intent
  supervision; $\delta\mathbf q_t^*$ denotes the correction to the nominal
  command. Overlaid configurations illustrate these distinctions.}
  \label{fig:haco-components}
\end{figure}

Choosing the action target is critical under sustained contact. The next
observed state $\mathbf q_{t+1}^{\rm obs}$ captures only the realized motion and
therefore omits commanded motion blocked by contact, as well as the
command--state discrepancy that sustains interaction forces. The nominal
teleoperation command $\mathbf q_t^{\rm nom}$ preserves the operator's motion
intent, but may induce excessive contact force because the operator does not
directly perceive the robot's contact load. We therefore use
$\mathbf q_t^{\rm cmp}$ as the learning target. It preserves task-directed
motion while incorporating the reference adjustment introduced by compliance
regulation. Figure~\ref{fig:haco-components} visualizes the resulting distinctions
among observed configurations, nominal commands, and compliant actions.

To expose the force-regulating content within the unified command, we define
the time-aligned command--state discrepancy
\begin{equation}
\Delta\mathbf q_t^{\rm ci}
  =\mathbf q_t^{\rm cmp}-\mathbf q_t^{\rm obs}.
\label{eq:compliance_intent}
\end{equation}
Under sustained contact, this time-aligned command--state discrepancy is
dominated by the contact-dependent reference offset introduced by compliance
regulation and thus captures the force-regulating component of the compliant
command. We therefore refer to $\Delta\mathbf q_t^{\rm ci}$ as the
\emph{compliant intent}. It is supervised as an auxiliary output rather than
executed as an independent residual.
Compliant actions and compliant intent are normalized using separate dataset
statistics.

We train HACo's action expert with conditional flow
matching~\cite{gr00tn1_2025}. The joint data target concatenates the
compliant-action and compliant-intent trajectories,
$\mathbf Y_1=[\mathbf A^{\rm cmp};\Delta\mathbf q^{\rm ci}]$, while the source
sample is drawn from $\mathbf Y_0\sim\mathcal N(\mathbf 0,\mathbf I)$. For a
flow time $s\sim\mathcal U[0,1]$, the interpolated sample and target velocity
are
\begin{equation}
\mathbf Y_s=(1-s)\mathbf Y_0+s\mathbf Y_1,
\qquad
\mathbf v^*=\mathbf Y_1-\mathbf Y_0.
\label{eq:flow_path}
\end{equation}
Conditioned on $\mathcal O_t$, the expert predicts
$\mathbf v_\theta(\mathbf Y_s,s,\mathcal O_t)$. Let $\mathcal P_{\rm act}$ and
$\mathcal P_{\rm ci}$ extract the compliant-action and compliant-intent
channels, respectively. We optimize
\begin{equation}
\begin{aligned}
\mathcal L={}&\mathbb E\!\left[
 \|\mathcal P_{\rm act}(\mathbf v_\theta-\mathbf v^*)\|_2^2\right]\\
&+\lambda_{\rm CIS}\mathbb E\!\left[
 \|\mathcal P_{\rm ci}(\mathbf v_\theta-\mathbf v^*)\|_2^2\right].
\end{aligned}
\label{eq:haco_objective}
\end{equation}
where $\lambda_{\rm CIS}=0.5$ and both terms are averaged over valid time steps
and channels. The first term learns the executable compliant trajectory, while
the second constitutes \emph{Compliant-Intent Supervision} (CIS). Optimizing
them jointly encourages the shared action representation to preserve the
force-regulating information encoded by the command--state discrepancy.

For asynchronous closed-loop execution, we adopt Real-Time Chunking
(RTC)~\cite{black2025rtc} with training-time prefix
conditioning~\cite{black2025trainingrtc}. During training, each action chunk is
conditioned on a sampled clean prefix, and both flow losses are evaluated only
on the remaining postfix. We set the maximum sampled prefix length to 12 steps.
At inference, each new chunk is conditioned on a 10-step prefix committed from
the preceding chunk, allowing the postfix to be generated from the latest
observation without interrupting execution. Only
$\widehat{\mathbf A}^{\rm cmp}$ is sent to the controllers; the predicted
compliant intent remains an auxiliary output.

\subsection{Kinematics-Aligned Haptic Perception}
\label{sec:method_haptic}

Fingertip tactile sensing and hand-joint torque feedback characterize
complementary aspects of physical interaction. At each instrumented fingertip,
a wrench captures the local contact load, while a deformation map captures the
spatial contact pattern. Joint torques contain a broader signature of external
loading transmitted through the articulated hand, but are spatially ambiguous.
HACo therefore aligns the two modalities according to the hand kinematics
before modeling interactions across digits.

Our platform measures 44 hand-joint torques and ten fingertip tactile streams.
For digit $i$ at time step $t$, let $\mathcal J_i$ denote the joints on its
kinematic chain and $\boldsymbol\tau_{\mathcal J_i,t-L+1:t}$ collect their $L$-step torque
histories. Let $\mathbf m_{i,t}^{\tau}$ and
$\mathbf m_{i,t}^{\rm tac}$ denote the corresponding validity indicators. A
torque encoder embeds the joint-torque histories and aggregates them into a
chain-level representation. In parallel, a tactile encoder combines the
fingertip's $L$-step wrench history $\mathbf W_i$ with its current deformation
map $\mathbf D_i$. The two representations are fused into one token for the
same digit before any cross-finger information exchange:
\begin{equation}
\begin{aligned}
\mathbf u_{i,t}^{\tau}
 &=E_\tau(\boldsymbol\tau_{\mathcal J_i,t-L+1:t},
          \mathbf m_{i,t}^{\tau}),\\
\mathbf u_{i,t}^{\rm tac}
 &=E_{\rm tac}(\mathbf W_{i,t-L+1:t},\mathbf D_{i,t},
               \mathbf m_{i,t}^{\rm tac}),\\
\bar{\mathbf p}_{i,t}
 &=E_{\rm fuse}([\mathbf u_{i,t}^{\tau};\mathbf u_{i,t}^{\rm tac}],
                 \mathbf m_{i,t}^{\tau},\mathbf m_{i,t}^{\rm tac}),\\
\mathbf p_{i,t}
 &=\bar{\mathbf p}_{i,t}+\mathbf e_{h_i}^{\rm hand}
                         +\mathbf e_{d_i}^{\rm digit},\\
\mathbf P_t
 &=E_{\rm ctx}(\{\mathbf p_{i,t}\}_{i=1}^{N_f};\mathbf m_t).
\end{aligned}
\label{eq:haptic_encoding}
\end{equation}
Here, $\mathbf e_{h_i}^{\rm hand}$ and $\mathbf e_{d_i}^{\rm digit}$ are
learned identity embeddings indexed by the hand $h_i$ and digit $d_i$ of token
$i$, respectively, and $\mathbf m_t$ is the resulting finger-validity mask.
This ordering binds the localized tactile evidence at each fingertip to the
load response of its own kinematic chain before $E_{\rm ctx}$ models coordinated
loading across digits. We use $L=9$ for the torque and wrench histories, while
the deformation branch receives only the current $240\!\times\!240$ map.
A cross-digit Transformer implements $E_{\rm ctx}$ over the $N_f=10$ finger
tokens while masking unavailable measurements. The resulting $\mathbf P_t$
serves as structured haptic memory for the Compliance Grounding Module.

\subsection{Compliance Grounding Module}
\label{sec:method_grounding}

Naively appending haptic features to the visual or action token sequence
does not explicitly align contact observations with the evolving action
representation and may disrupt the cross-modal dependencies acquired during
pretraining. We therefore introduce the \emph{Compliance Grounding Module}
(CGM), which allows the current action features to query the structured haptic
tokens $\mathbf P_t$ through gated haptic cross-attention. For action features
$\mathbf Z^l$ at layer $l$, CGM computes
\begin{equation}
\begin{aligned}
\mathbf C_t^l
 &=\operatorname{Attn}\!\left(
   \operatorname{AdaLN}(\mathbf Z^l,s),
   \mathbf P_t,\mathbf P_t;\mathbf m_t\right),\\
\widetilde{\mathbf Z}^{l}
 &=\mathbf Z^l+\tanh(\alpha_l)\mathbf C_t^l,
\end{aligned}
\label{eq:compliance_grounding}
\end{equation}
where $\mathbf m_t$ is the haptic availability mask and $\alpha_l$ is a
learnable residual gate. The gate allows each adapted layer to regulate its
reliance on haptic evidence. We adopt zero initialization for the residual
gate, ensuring that CGM begins as an identity mapping and preserves the
pretrained computation at the outset of fine-tuning.

\begin{figure}[!b]
  \centering
  \includegraphics[width=0.98\columnwidth]{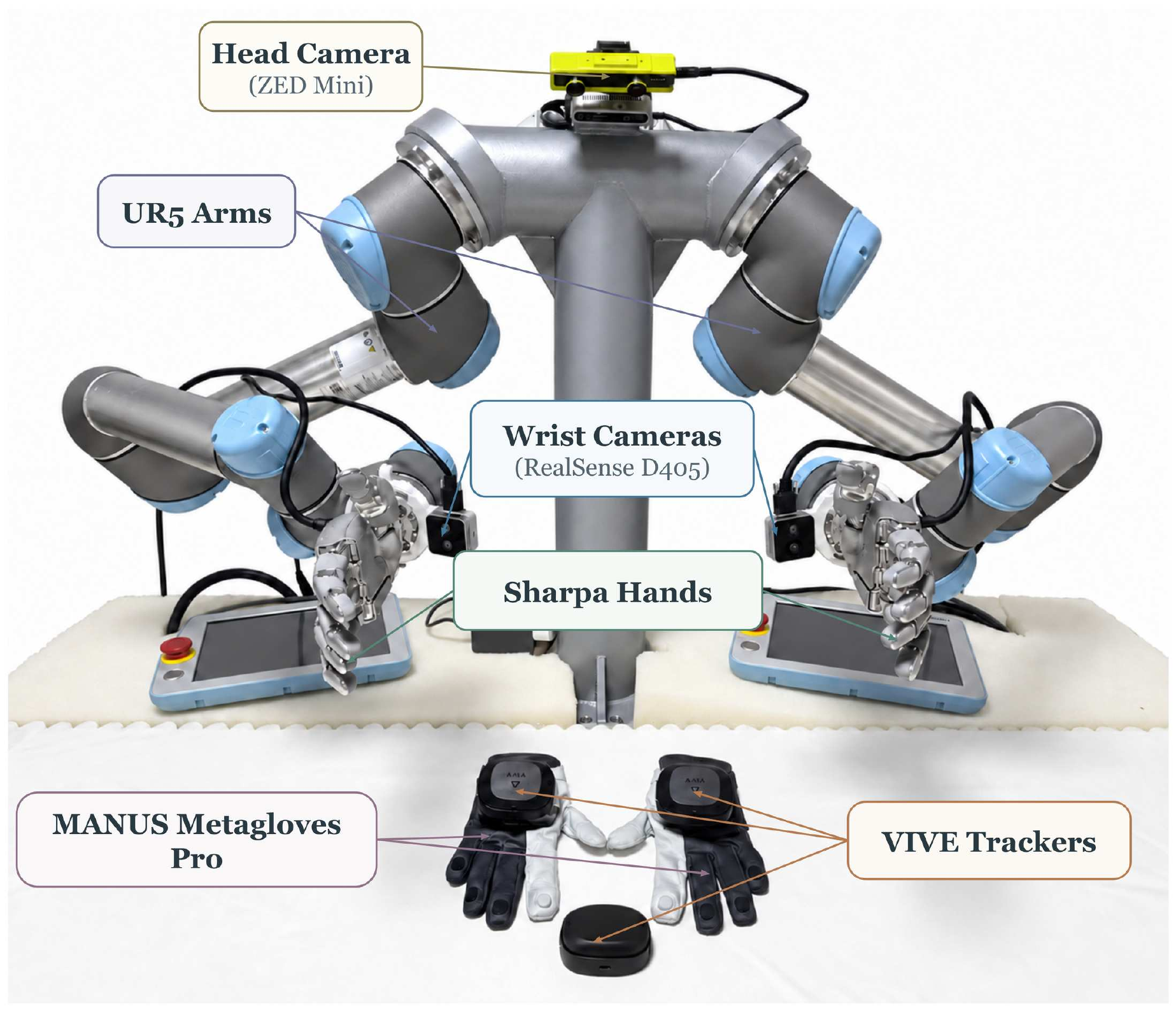}
  \caption{\textbf{Teleoperation and experimental platform.} The operator uses
  MANUS Metagloves Pro and VIVE Trackers to capture finger articulation and wrist pose,
  which are retargeted to the Sharpa hands and UR5 end-effectors. A ZED Mini and
  two wrist-mounted D405 cameras provide visual observations; robot state,
  tactile feedback, and joint torques are recorded synchronously.}
  \label{fig:teleoperation-platform}
\end{figure}

\begin{table*}[t]
  \caption{\textbf{Policy comparison on the real-world force-sensitive
  dexterous manipulation benchmark.} Entries report successful rollouts out
  of 20; Mean is the macro-average success rate across five tasks.}
  \label{tab:main-results}
  \centering
  \setlength{\tabcolsep}{7.0pt}
  \small
  \begin{tabular}{@{}lcccccc@{}}
    \toprule
    Method & Insert poker cards & Open book & Draw on balloon & Unscrew cap & Squeeze toothpaste & Mean \\
    \midrule
    GR00T~\cite{gr00tn1_2025} & 3/20 & 0/20 & 1/20 & 7/20 & 4/20 & 15\% \\
    GR00T + Tactile & 5/20 & 2/20 & 1/20 & 9/20 & 5/20 & 22\% \\
    ViTacFormer~\cite{heng2025vitacformer} & 0/20 & 1/20 & 0/20 & 2/20 & 1/20 & 4\% \\
    T-Rex~\cite{niu2026trex} & 4/20 & 6/20 & 2/20 & 12/20 & 11/20 & 35\% \\
    \midrule
    \textbf{HACo} & \textbf{18/20} & \textbf{17/20} & \textbf{14/20} & \textbf{19/20} & \textbf{15/20} & \textbf{83\%} \\
    \bottomrule
  \end{tabular}
\end{table*}

\begin{table*}[t]
  \caption{\textbf{Ablation results.} Entries report successful rollouts out
  of 20; parenthesized values denote the decrease in mean success rate from
  full HACo.}
  \label{tab:ablations}
  \centering
  \setlength{\tabcolsep}{4.2pt}
  \footnotesize
  \begin{tabular}{@{}lcccccc@{}}
    \toprule
    Configuration & Insert poker cards & Open book & Draw on balloon & Unscrew cap & Squeeze toothpaste & Mean \\
    \midrule
    \textbf{HACo} & \textbf{18/20} & \textbf{17/20} & \textbf{14/20} & \textbf{19/20} & \textbf{15/20} & \textbf{83\%} \\
    \midrule
    \rowcolor{gray!12}\multicolumn{7}{@{}l}{\textit{Haptic perception}} \\
    \quad w/o Haptic Feedback & 7/20 & 2/20 & 3/20 & 8/20 & 7/20 & 27\%\,\abldrop{56} \\
    \quad w/o Tactile Feedback & 8/20 & 5/20 & 7/20 & 13/20 & 12/20 & 45\%\,\abldrop{38} \\
    \quad w/o Torque Feedback & 15/20 & 16/20 & 11/20 & 14/20 & 12/20 & 68\%\,\abldrop{15} \\
    \quad w/o Coupled Encoding & 17/20 & 14/20 & 12/20 & 13/20 & 14/20 & 70\%\,\abldrop{13} \\
    \addlinespace[2pt]
    \rowcolor{gray!12}\multicolumn{7}{@{}l}{\textit{Compliance learning (cumulative removal)}} \\
    \quad w/o Compliant-Intent Supervision & 15/20 & 15/20 & 13/20 & 16/20 & 14/20 & 73\%\,\abldrop{10} \\
    \qquad Nominal Action, w/o CIS & 12/20 & 11/20 & 9/20 & 14/20 & 13/20 & 59\%\,\abldrop{24} \\
    \addlinespace[2pt]
    \rowcolor{gray!12}\multicolumn{7}{@{}l}{\textit{Compliance grounding}} \\
    \quad Visuo--Haptic Fusion & 10/20 & 6/20 & 7/20 & 10/20 & 8/20 & 41\%\,\abldrop{42} \\
    \quad Action-Suffix Fusion & 14/20 & 9/20 & 6/20 & 11/20 & 7/20 & 47\%\,\abldrop{36} \\
    \quad CGM w/o Gate & 14/20 & 15/20 & 11/20 & 16/20 & 10/20 & 66\%\,\abldrop{17} \\
    \addlinespace[2pt]
    \rowcolor{gray!12}\multicolumn{7}{@{}l}{\textit{Visual observation}} \\
    \quad w/o Wrist Cameras & 13/20 & 15/20 & 14/20 & 18/20 & 16/20 & 76\%\,\abldrop{7} \\
    \bottomrule
  \end{tabular}
\end{table*}

\begin{figure*}[t]
  \centering
  \includegraphics[width=\textwidth]{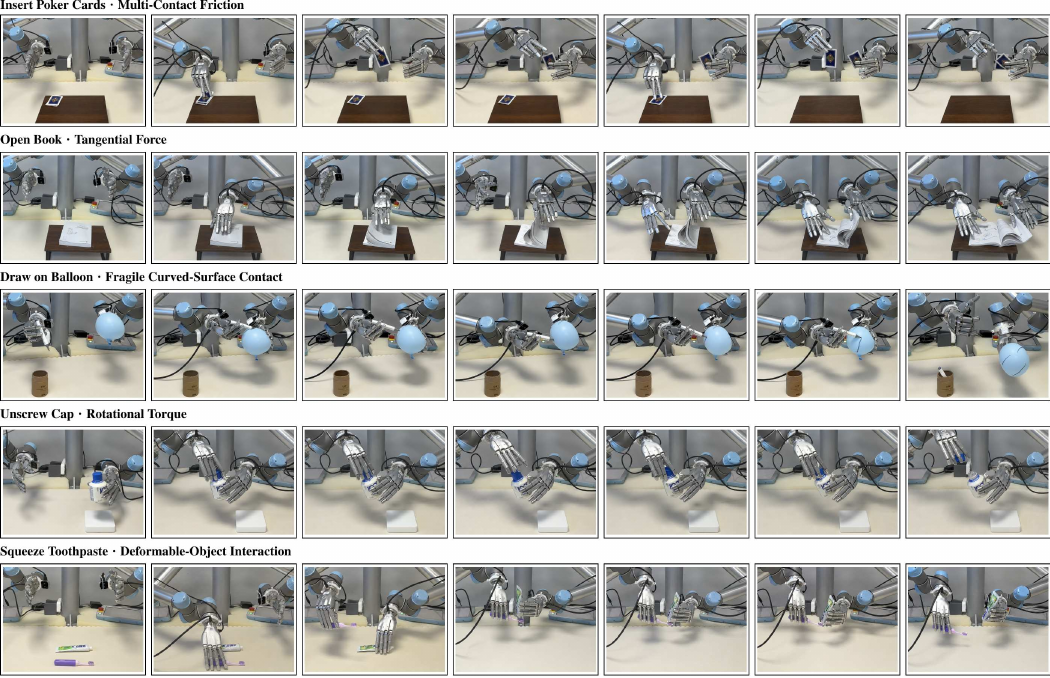}
  \caption{\textbf{Representative HACo rollouts on the real-world dexterous
  force benchmark.} Each row shows seven temporally ordered frames of one
  execution, progressing from contact establishment to task completion.
  Better viewed in videos.
  }
  \label{fig:benchmark-tasks}
\end{figure*}

\section{Experiments}
\label{sec:experiments}

Our experiments address four questions: (Q1) Does HACo outperform existing
policies on force-sensitive dexterous tasks? (Q2) How do tactile and joint-torque
feedback contribute, and are they complementary? (Q3) What are the respective
roles of compliant-action targets and CIS? (Q4) How should haptic features be
incorporated into the pretrained action model? We also evaluate wrist-camera
observations. Each task--method pair is evaluated over 20 physical rollouts,
with results averaged across the five tasks.

\subsection{Experimental Setup}
\label{sec:experimental-setup}

\paragraph{Platform and observations}
Experiments are conducted on a dual-arm system comprising two UR5 manipulators
equipped with Sharpa dexterous hands. The Sharpa hands provide integrated
fingertip tactile sensing and hand joint-torque feedback. Visual observations
are captured by a ZED Mini egocentric camera and two wrist-mounted Intel
RealSense D405 cameras. HACo receives the three RGB views, robot kinematic
state, fingertip tactile feedback, and joint-torque feedback, and replans
40-step action chunks in closed loop. The complete physical system and
teleoperation interface are shown in Figure~\ref{fig:teleoperation-platform}.
MANUS Metagloves Pro
capture finger articulation for retargeting to the Sharpa hands, while VIVE
Trackers provide wrist poses for commanding the UR5 end-effectors.

\paragraph{Training details}
For post-training, we collect 100 demonstrations per task. All policies use the
complete task-specific dataset and are optimized for 30k steps on four NVIDIA
H100 GPUs with a per-GPU batch size of 12
(global batch size 48), and seed 42. HACo, GR00T, and GR00T + Tactile are
initialized from GR00T N1.7 and optimized with AdamW using a learning rate of
$2\!\times\!10^{-5}$, weight decay $10^{-5}$, and a 5\% warmup. GR00T predicts
the nominal action from vision and robot state. GR00T + Tactile retains the
same architecture and optimization settings, but concatenates the normalized
fingertip tactile feature vector with the robot state before the original state
projector. HACo instead predicts the compliant action together with
compliant-intent supervision. T-Rex is initialized from its released
mid-training checkpoint and optimized with AdamW at a learning rate of
$10^{-4}$, zero weight decay, and a 5\% warmup. ViTacFormer uses its
ImageNet-pretrained ResNet-18 initialization and AdamW with learning rates of
$10^{-4}$ for the policy and $10^{-5}$ for the visual backbone, weight decay
$10^{-4}$, and 1k warmup steps. The HACo ablations use the same training
configuration as the full model.

\paragraph{Evaluation protocol}
We conduct 20 physical rollouts for every task--method pair. Each rollout
is counted as successful when the task is completed. We report the number of
successful rollouts out of 20 for each task, and the final column gives the
macro average of the five task success rates.

\subsection{Real-World Dexterous Force Benchmark}
\label{sec:force-benchmark}

Most prior work evaluates tasks centered on motion outcomes rather than force
regulation. To explicitly evaluate force-aware manipulation, we consider five
multi-stage bimanual tasks spanning distinct force-sensitive interaction
regimes. \textbf{Insert poker cards} tests
multi-contact friction by requiring one playing card to be aligned and slid
into an occupied grasp without bending or dropping either card. \textbf{Open
book} probes precise tangential-force control: the thumb must accurately
regulate both the direction and magnitude of tangential force to separate an
interior section of pages, turn it across the spine, and hold the book open
without damage.
\textbf{Draw on balloon} requires drawing a smiley face on a fragile curved
surface while avoiding slip, excessive deformation, or rupture. \textbf{Unscrew
cap} tests rotational torque as one hand stabilizes the bottle and the other
removes the cap without drops or spills. \textbf{Squeeze toothpaste} tests
controlled compression of a deformable container to dispense toothpaste onto
toothbrush bristles. These
regimes emphasize different interaction demands, but each task can involve
several contact mechanisms. Together, these tasks
require a policy to complete the intended motion while regulating the physical
interaction that makes completion possible. Representative executions across
all five tasks are shown in Figure~\ref{fig:benchmark-tasks}.

\subsection{Comparison with Existing Policies}
\label{sec:main-comparison}

For the overall policy comparison, Table~\ref{tab:main-results} reports the
results under the evaluation protocol above. HACo attains an 83\% mean success
rate, exceeding the strongest baseline, T-Rex (35\%), by 48\% and GR00T
(15\%) by 68\%. The advantage holds across all five tasks.

The baseline results expose distinct architectural limitations. GR00T +
Tactile improves only modestly to 22\%: concatenating tactile features with
robot state provides no explicit alignment between contact evidence and the
evolving action representation.
T-Rex~\cite{niu2026trex} averages 57.5\% on the two tasks dominated by sustained
rotational or compressive loading, versus 20\% on the three tasks with more
abrupt force transitions. Its tactile-only refinement adds inference latency and reuses
cached visual context, potentially limiting responsiveness to rapid
visual--force changes. ViTacFormer~\cite{heng2025vitacformer} reaches 4\%; its
compact ACT policy lacks the large-scale visuomotor pretraining of the VLA
baselines, which may limit its capacity to represent the complex bimanual
motions in our tasks.
Together, these results favor action-aligned haptic conditioning built on a
pretrained action model.

\subsection{Ablation Studies}
\label{sec:ablations}

Table~\ref{tab:ablations} ablates haptic perception, compliance learning,
compliance grounding, and wrist-camera observations.

\paragraph{Haptic perception}
Removing all haptic feedback lowers the mean score from 83\% to 27\%, confirming
that physical observations are essential for these force-sensitive tasks.
Fingertip tactile feedback alone reaches 68\%, compared with 45\% for joint
torque alone, indicating that local contact measurements provide the dominant
haptic cue. Nevertheless, adding joint torque to tactile feedback improves
every task and raises the cap-removal success rate from 70\% to 95\%. This gain is consistent
with joint torque capturing loads transmitted through the articulated hand
during coordinated rotation. The two modalities therefore provide
complementary rather than redundant information. Coupled encoding further
improves the mean success rate from 70\% to 83\%, suggesting that preserving
kinematic correspondence helps coordinate contact loads across multiple digits.

\paragraph{Compliance learning}
Removing compliant-intent supervision (CIS) while retaining the compliant
action target reduces the mean score from 83\% to 73\%. Replacing the compliant
action with the nominal command in the same no-CIS setting further lowers it to
59\%. These 10\% and additional 14\% drops show that the action target contributes
more, while CIS remains complementary. We compute mean fingertip force over
contact-active samples. HACo reduces it by 19\% relative to the nominal-action
variant, consistent with more compliant interaction.

\paragraph{Compliance grounding}
Visuo--Haptic Fusion and Action-Suffix Fusion reach only 41\% and 47\%,
respectively, compared with 66\% for ungated haptic cross-attention and 83\%
for the complete CGM. Directly inserting haptic features into the
visual memory or action sequence perturbs the structure learned during
visuomotor pretraining and provides no explicit action-dependent access to
contact evidence. Dedicated cross-attention instead aligns haptic observations
with the evolving action representation. The zero-initialized gate further
preserves the pretrained computation at the start of adaptation and allows the
model to regulate its reliance on noisy or intermittently informative haptic
signals.

\paragraph{Wrist-camera observations}
Although wrist cameras are common in dexterous policy learning, removing both
reduces the mean success rate by only 7\%, from 83\% to 76\%.
This modest drop suggests that haptic feedback can partly compensate for
contact states hidden by visual occlusion, while wrist views remain
complementary.

\section{Conclusion}
\label{sec:conclusion}

Force-aware dexterous manipulation requires both physical feedback and action
targets that encode appropriate contact responses. \method learns
controller-executable compliant actions from compliance-regulated teleoperation,
preserves their command--state discrepancy through compliant-intent supervision,
and grounds pretrained action generation in kinematics-aligned tactile and
joint-torque feedback through gated haptic cross-attention. Across five real-world tasks,
HACo achieves an 83\% mean success rate, 48\% above the
strongest evaluated baseline. Ablations validate the complementary haptic
signals and action-aligned compliance learning, establishing a practical route
to closed-loop, force-aware dexterity.

Despite these advances, several challenges remain open. First, force-feedback
teleoperation is needed to capture intentional
human force strategies rather than correcting force-blind demonstrations after
the fact. Second, scalable dexterous learning must extend beyond teleoperation,
which bounds policies to demonstrable motions; egocentric human data and
reinforcement learning offer complementary routes to skills that cannot be
readily retargeted. Third, dexterous hardware should progress toward dense
whole-hand tactile coverage, making palm and phalange contacts directly
observable rather than indirectly inferred from joint torque.

\section*{Acknowledgment}
We would like to thank BAAI for providing the experimental equipment and
computational resources, as well as the teleoperators for collecting the
demonstrations. We thank Yankai Fu for insightful discussions and valuable
suggestions on the design of the demonstration tasks. We also thank our
colleagues at OpenDriveLab for their valuable insights and constructive
discussions.

\bibliographystyle{IEEEtran}
\bibliography{references}

\end{document}